\documentclass[default,iicol]{sn-jnl}

\usepackage{dsfont}
\usepackage{multirow}
\usepackage{bm}
\usepackage{caption}
\usepackage{subcaption}

\usepackage{graphicx}%
\usepackage{multirow}%
\usepackage{amsmath,amssymb,amsfonts}%
\usepackage{amsthm}%
\usepackage{mathrsfs}%
\usepackage[title]{appendix}%
\usepackage{xcolor}%
\usepackage{textcomp}%
\usepackage{manyfoot}%
\usepackage{booktabs}%
\usepackage{algorithm}%
\usepackage{algorithmicx}%
\usepackage{algpseudocode}%
\usepackage{listings}%

\theoremstyle{thmstyleone}%
\theoremstyle{thmstyletwo}%

\theoremstyle{thmstylethree}%

\begin{document}

\title[Article Title]{Graph data augmentation with Gromow-Wasserstein Barycenters}


\author*[1]{\fnm{Andrea} \sur{Ponti}}\email{a.ponti5@campus.unimib.it}

\affil*[1]{\orgdiv{Department of Economics, Management and Statistics}, \orgname{University of Milano-Bicocca}, \orgaddress{\city{Milan}, \postcode{20126}, \country{Italy}}}


\abstract{Graphs are ubiquitous in various fields, and deep learning methods have been successful applied in graph classification tasks. However, building large and diverse graph datasets for training can be expensive. While augmentation techniques exist for structured data like images or numerical data, the augmentation of graph data remains challenging. This is primarily due to the complex and non-Euclidean nature of graph data.  In this paper, it has been proposed a novel augmentation strategy for graphs that operates in a non-Euclidean space. This approach leverages graphon estimation, which models the generative mechanism of networks sequences. Computational results demonstrate the effectiveness of the proposed augmentation framework in improving the performance of graph classification models. Additionally, using a non-Euclidean distance, specifically the Gromow-Wasserstein distance, results in better approximations of the graphon. This framework also provides a means to validate different graphon estimation approaches, particularly in real-world scenarios where the true graphon is unknown.}




\maketitle

\section{Introduction}\label{sec1}
Networks are a powerful tool used in many fields such as social sciences or biology, and they are becoming larger and more complex. Deep Learning methods, like Graph Neural Networks and Graph Convolutional Networks, have been successfully applied to perform classification on graphs. Building graph datasets to train neural methods can be very expensive in real-word applications. When dealing with structured data, as numerical data or images, many augmentation techniques exist to face this problem. In the context of graph classification this is still challenging. Existing graph data augmentation strategies typically aim to augment graphs at a \textit{within-graph} level by either modifying edges or nodes in an individual graph, which does not enable information exchange between different instances. The \textit{between-graph} augmentation methods are still under-explored. This is mainly due to the complex and non-Euclidean structure of graph data \cite{zhao2022graph}. Such complexity hinders the direct application of traditional augmentation techniques used in image, video, or text data. \\
One way to study large and complex networks is by modelling their generative mechanism in order to build statistical graph models. Graphons (short for graph functions), first introduced in \cite{borgs2008convergent, lovasz2012large}, are the limiting objects for sequences of large, finite graphs with respect to the so-called cut metric. Formally, a graphon is a measurable symmetric function $W: \Omega \times \Omega \to [0,1]$, where $\Omega$ is a measurable space, that represents a discrete network as an infinite-dimensional analytic object. A graphon can be used to generate arbitrarily sized graphs by first sampling $N$ nodes from a uniform distribution, $v_i \sim \text{Uniform}(\Omega)$, and then generating an adjacency matrix $A = [a_{ij}]$. Each element $a_{ij} \in \{0, 1\}$ of the adjacency matrix is sampled from a Bernoulli distribution determined by the graphon $a_{ij} \sim \text{Bernoulli}(W(v_i, v_j))$. Accordingly, the generated graph is $\mathcal{G}=(V, E)$ where $V=\left\{v_1, ..., v_N\right\}$ and $E = \left\{(v_i, v_j) | a_{ij} = 1\right\}$. Graphons arise naturally in many fields like random networks, quasi-random graphs, property testing of large graphs and extremal graph theory \cite{lovasz2006limits, janson2011quasi, han2021quasi}. They find applications in fields like signal theory, where they have been used to overcome the limitations of Graph Signal Processing in the case of graphs with large number of nodes or dynamic topologies \cite{ruiz2021graphon}. Graphon have been also used in Graph Neural Networks (GNNs) to analyze the transferability of GNNs across graphs \cite{ruiz2020graphon}, their stability \cite{ruiz2021graphonstability}, and to mitigate the over-smoothing issue via resampling from a graphon estimate obtained from the underlying network data \cite{hu2021training}. \\
Many methods to learn a graphon from an observed sequence of graph are based on the weak regularity lemma, which states that an arbitrary graphon can be well approximated by a two-dimensional step function. Existing methods to learn a step function typically leverage on stochastic block models \cite{airoldi2013stochastic, chan2014consistent, channarond2012classification} or low-rank approximation of the observed graphs \cite{keshavan2010matrix, chatterjee2015matrix, frieze1999quick}.
Most of the methods proposed in the literature requires the observed graphs to be generated by a single graphon and to be aligned, i.e., graphs with comparable size and with a known correspondence between nodes. In real world graphs, this is often not the case. Traditional learning methods involve an aligning procedure to match the observed graph using some heuristics that often produces failures, resulting in wrongly-aligned graphs and in poor estimation of the underlying graphon. To overcome these limitations, the authors in \cite{xu2021learning} proposed an approach that leverages the permutation-invariance of the Gromov-Wasserstein (GW) distance, implicitly including the graph aligning in the estimation phase. This approach will be further explored in Section \ref{sec:graphon_wst}. \\
The purpose of this paper is to propose an augmentation strategy for graphs that works in a non-Euclidean space. Experimental results, in a graph classification context, show an improvement in performance when using the augmented dataset. Furthermore, results show that using a non-Euclidean distance results in a better approximation of the graphon. The proposed framework can also be used to validate different graphon estimation approaches, specifically in real-word problems when the true graphon is unknown.

\section{Graphons as Wasserstein Barycenters} \label{sec:graphon_wst}
Consider a set of graphs $\{ \mathcal{G}_m \}_{m=1}^M$. A graph with nodes $V=\{1,...,N\}$ and an adjacency matrix $A=[a_{n,n'}]$ can be represented as a step function $G_{\mathcal{P}}$ over $N$ equitable partitions of $\Omega$, $\mathcal{P} = \{ \mathcal{P}_n \}_{n=1}^N$, defined as
\begin{equation}
    G_{\mathcal{P}}(x,y) = \frac{1}{N^2} \sum_{n,n'=1}^N a_{n,n'} \mathds{1}_{\mathcal{P}_n \times \mathcal{P}_{n'}}(x,y).
\end{equation}
where $a_{n,n'}\in [0,1]$ and the indicator function $\mathds{1}_{\mathcal{P}_n \times \mathcal{P}_{n'}}(x,y)$ takes value 1 if $(x,y) \in \mathcal{P}_n \times \mathcal{P}_{n'}$ and 0 otherwise. If the position of nodes is known, it is possible to derive an isomorphism of the graph $\hat{\mathcal{G}}$ and to obtain an ``oracle'' step function $\hat{G}_\mathcal{P}$. Using such isomorphism over the set of all graphs, the result is a set of aligned graphs $\{ \hat{\mathcal{G}}_m \}_{m=1}^M$ and consequently, a set of oracle step functions $\{ \hat{G}_{m, \mathcal{P}^m} \}_{m=1}^M$, where the number of partitions $|\mathcal{P}^m|$ is equal to the number of nodes in $\hat{\mathcal{G}}_m$. Then, an oracle estimator of the graphon $W$ that generated the set of graphs $\{ \mathcal{G}_m \}_{m=1}^M$ can be obtained as
\begin{equation}
    W_O = \frac{1}{M} \sum_{m=1}^M \hat{G}_{m, \mathcal{P}^m}
\end{equation}
The authors in \cite{xu2021learning} have shown that the oracle estimator provides a consistent estimation of $W$ indeed, for every $W \in \mathcal{W}$, the following inequality is true
\begin{equation}
\delta_\square (W, W_O) \leq \frac{C}{\min_m|\mathcal{P}^m|}
\label{eq:consistency}
\end{equation}
with $C$ a constant. Using Equation \ref{eq:consistency} and the triangle inequality of the cut-distance, in \cite{xu2021learning} it has been shown that  the estimation of a graphon can be done minimizing an upper bound of the cut-distance by solving the following optimization problem:
\begin{equation}
    \min_{W_\mathcal{P}} \frac{1}{M} \sum_{m=1}^M \delta_1(G_{m,\mathcal{P}^m}, W_\mathcal{P}).
\end{equation}
To solve this problem, the authors in \cite{xu2021learning} derived a computationally-efficient alternative of $\delta_1$ based on its equivalence with the 1-order GW distance. Consider two step functions $W_{1,\mathcal{P}}$ and $W_{2,\mathcal{Q}}$ defined on the probability spaces $(\Omega_1,\mu_1)$ and $(\Omega_2,\mu_2)$, respectively, which have equitable partitions $\mathcal{P} = \{ \mathcal{P}_i \}_{i=1}^I$ and $\mathcal{P} = \{ \mathcal{Q}_j \}_{j=1}^J$. Then, the $\delta_1$ distance between the two step functions can be rewritten as the 1-order GW distance:
\begin{align}
    d_{gw, 1}(\textbf{W}_1, \textbf{W}_2) & = \\
    \min_{\textbf{T} \in \Pi(\boldsymbol{\mu}_1, \boldsymbol{\mu}_2)} & \sum_{i,i',j,j'} |w_{1,ij} - w_{2,i'j'}| T_{ii'} T_{jj'} \nonumber
\end{align}
where $\textbf{W}_1=[w_{1,ij}] \in [0, 1]^{I \times I}$ and $\textbf{W}_2=[w_{2,i'j'}] \in [0, 1]^{J \times J}$ represent the step functions in the matrix form; vectors $\boldsymbol{\mu}_1 = [\mu_{1,i}]$ and $\boldsymbol{\mu}_2 = [\mu_{2,j}]$ represent the probability measures $\mu_1$ and $\mu_2$; the matrix $\textbf{T}$ is the coupling between $\boldsymbol{\mu}_1$ and $\boldsymbol{\mu}_2$, and is also called transport matrix. Similarly it is possible to define the squared 2-order GW distance as
\begin{align}
    d_{gw, 2}^2(\textbf{W}_1, \textbf{W}_2) & \\
    = \min_{\textbf{T} \in \Pi(\boldsymbol{\mu}_1, \boldsymbol{\mu}_2)} & \sum_{i,i',j,j'} |w_{1,ij} - w_{2,i'j'}|^2 T_{ii'} T_{jj'}. \nonumber
    \label{eq:gw2}
\end{align}
This is usually smother and consequently easier to optimize. Then, the learning problem becomes the estimation of a GW barycenter of the observed graphs
\begin{equation}
    \min_{\textbf{W} \in [0,1]^{K \times K}} \frac{1}{M} \sum_{m=1}^M d_{gw,2}^2(\textbf{A}_m, \textbf{W})
    \label{eq:gwb}
\end{equation}
where $\textbf{A}_m$ is the adjacency matrix of the graph $\mathcal{G}_m$ and $\textbf{W} = [w_{kk'}]$ is the matrix representation of the step function $W_\mathcal{P}$.

\section{Graphon for data augmentation}\label{sec3}
In the view of graphons as generative models of graph objects they become particularly useful to augment graph datasets. This idea has been recently explored in \cite{han2022g}, where the authors derive a mixup graphon from the graphons of each class and use the derived mixup graphon to generate synthetic graphs with uncertain labels (i.e., the probability of each class is uniform). They show that adding these synthetic graphs to the initial dataset improves the generalization ability of GNNs and stabilize the model training. In the current paper, a slightly different approach has been proposed. Consider the supervised problem of classify a set of $M$ graphs $\bm{\mathcal{G}} = \{ \mathcal{G}_m \}_{m=1}^M$ into $N$ classes. Denote with $\bm{\mathcal{G}}_i$ the set of graphs belonging to class $i$. Then, for each class $i$, a graphon $\textbf{W}_i$ can be estimated from the sequence of graphs $\bm{\mathcal{G}}_i$. Finally, new graphs can be sampled from $\textbf{W}_i$ to augment the initial dataset. \\
In the computational results, reported in the following section, the datasets have been divided into train and test set; the graphons of each class are learned considering only the graphs in the train set. Code and data used are available on GitHub\footnote{\url{https://github.com/andreaponti5/graphon-data-augmentation}}.

\section{Experiments and Results}\label{sec4}
In the experiments three datasets have been considered to analyze the performance of the proposed augmentation framework. LFR and IMDB are two binary classification datasets of 1600 and 1000 graphs, respectively, while ENZYMES contains 600 graphs evenly distributed into 6 classes. In addition, five different methods to estimate graphons have been tested. SAS, SBA and LG are based on stochastic block models while MC is based on low-rank approximation. GB and SGB use the Gromow-Wasserstein barycenter to learn the graphons; in particular GB is reported in Equation \ref{eq:gwb} and SGB is a smoothed version of GB.
\begin{figure}[htp]

\subfloat[LFR dataset.]{%
  \includegraphics[clip,width=\columnwidth]{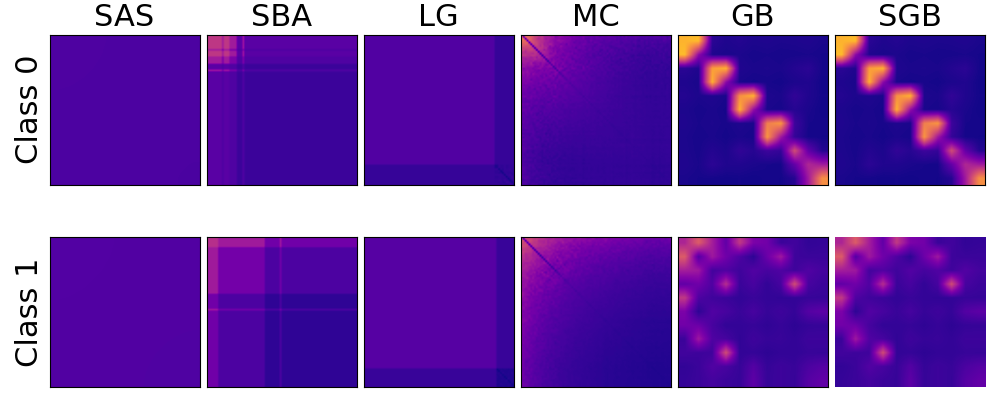}%
}

\subfloat[IMDB dataset.]{%
  \includegraphics[clip,width=\columnwidth]{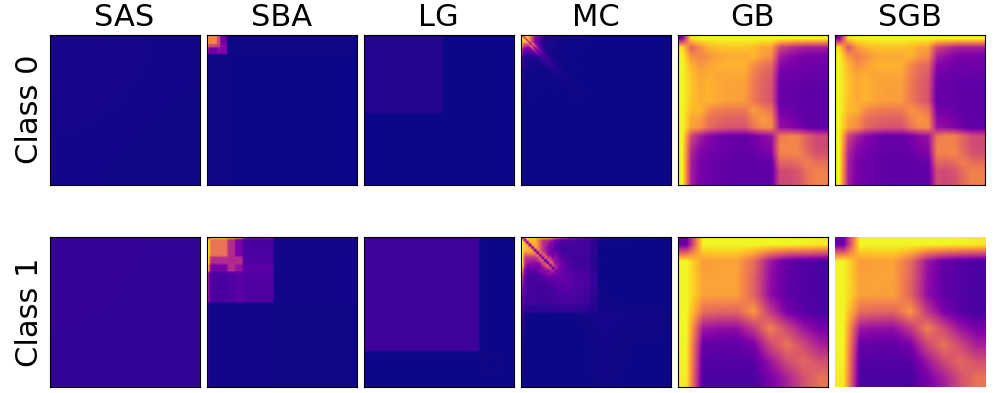}%
}

\subfloat[ENZYMES dataset.]{%
  \includegraphics[clip,width=\columnwidth]{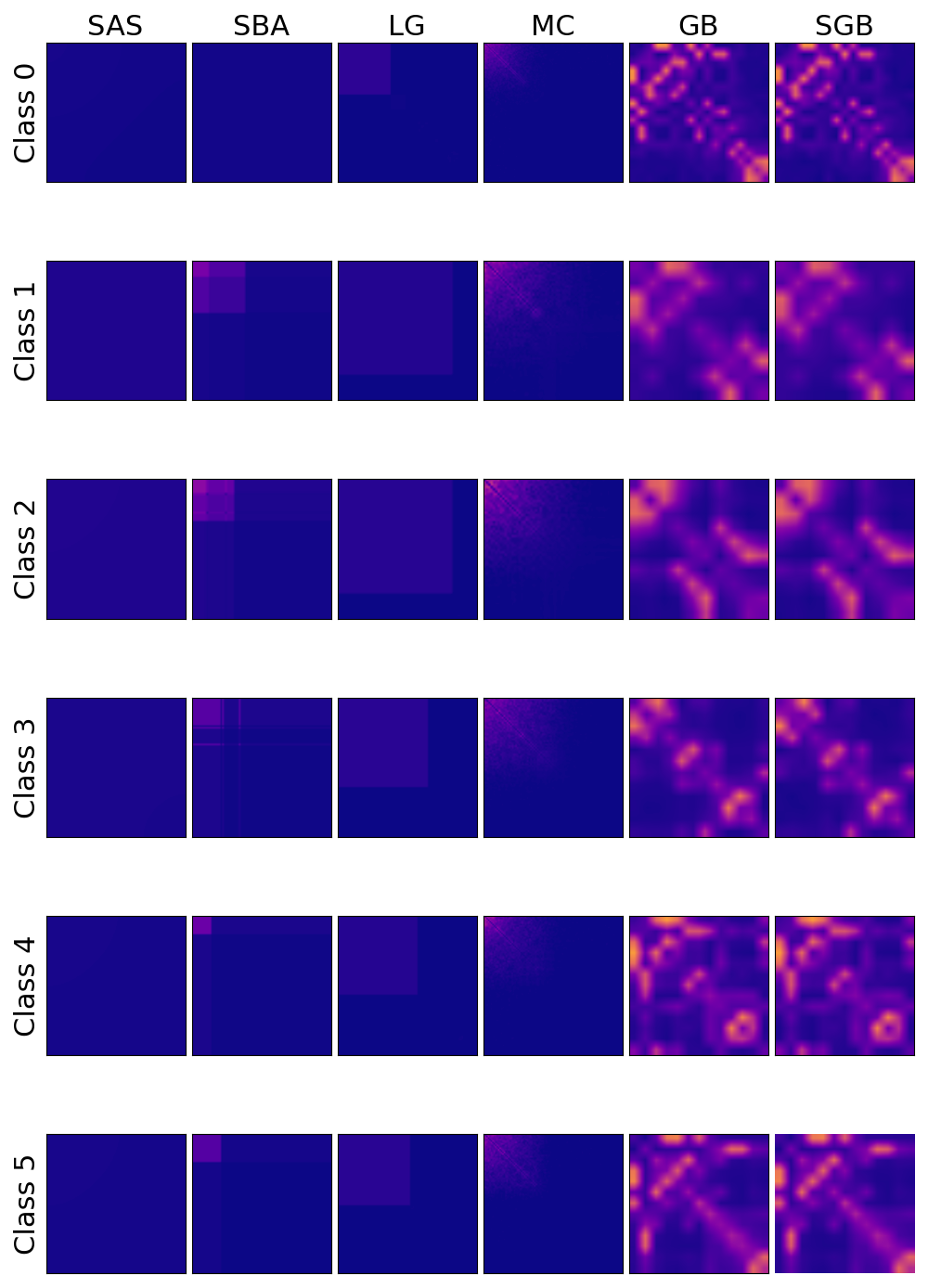}%
}

\caption{Visual representation of the graphons estimated using different methods on the three datasets. Lighter colors means higher probability of connection.}
\label{fig:graphons}
\end{figure}\\
First, it is important to understand whether graphons are able to discriminate the different classes in the dataset. Fig. \ref{fig:graphons} displays a visual representation of the graphons estimated with different methods of the classes of the three datasets. \\
The Wasserstein based estimations show a clear distinction for the two classes and give some insight on the structure of graphs belonging to the different classes. For instance, considering LFR, the graphs of class 0 appears to be composed by many connected community, while graphs in class 1 are more likely to be a single community with higher nodes degree. The other four baseline estimation methods do not show a marked difference between the two classes.\\
Then, to evaluate the quality of the generated graphs, they have been used to augment the initial dataset as explained in the previous section. Table \ref{tab:res} reports the percentage of classification accuracy improvement by augmenting the initial dataset. 
\begin{table}[htp]
\caption{Percentage of accuracy improvement augmenting the dataset by sampling from graphons estimated using different methods. The datasets have been augmented by 1\%, 5\%, 10\% and 25\% of the number of instances in the train set. The base accuracy (percentage) is shown in square brackets next to the dataset name}
\begin{tabular}{lccccccc}
\hline
                  & \% & SAS & SBA & LG & MC & GB & SGB \\ \hline \hline
\multirow{4}{*}{\rotatebox[origin=c]{90}{LFR[94]}}    & 1               & 0.2     & 0.2     &  0.2   &  0.2   & 0.2    &   0.2    \\
                         & 5               &   0.2   &  0.2    &  0.2   & 0.2    &  \textbf{0.4}    &   0.2    \\
                         & 10              &   0.2   &  0.2    &  0.2   & 0.4    &  \textbf{0.6}    &  0.4    \\
                         & 25              &   \textbf{0.6}   &  0.2    &  0.4   &  0.2   &  0.2    &  0.4    \\ \hline
\multirow{4}{*}{\rotatebox[origin=c]{90}{IMDB[54]}}     & 1               &  0.7    &   0.7   &  \textbf{1.0}   &  0.7   &  0.7    &  0.7     \\
                         & 5               &  \textbf{1.7}    &  1.3    &  0.0   &  0.0   &  0.7    &   \textbf{1.7}    \\
                         & 10              &  -1.0    &  0.0    &  0.7   &  1.3   &  \textbf{1.7}    &   0.0    \\
                         & 25              &  -1.3    &  -1.0    &  -3.3   &  0.0   &  \textbf{0.3}    &  \textbf{0.3}     \\ \hline
\multirow{4}{*}{\rotatebox[origin=c]{90}{ENZ[17]}} & 1               & 5.6     & 2.2     & 5.6    &  5.6   &  \textbf{6.1}    &  5.6     \\
                         & 5               &  2.2    &  \textbf{2.8}    &  0.0   &  -2.2   &   1.1   &  1.1     \\
                         & 10              &  2.8    &  \textbf{3.9}    &  1.7   &  3.3   &  2.8    &   2.8    \\
                         & 25              &  2.8    &  0.0    &  1.1   &  1.1   &  \textbf{2.2}    &   1.1    \\ \hline
\label{tab:res}
\end{tabular}
\end{table}\\
In particular, graph2vec \cite{narayanan2017graph2vec} has been used to embed the graphs in a vector space (considering the nodes degree as features) and a Multi-Layer Perceptron to perform classification on graphs. In the case of LFR, where the classes are clearly separated and the classification accuracy of the model trained with no augmentation is already high, it is possible to obtain a little improvement adding synthetic graphs to the train set. This improvement increases when the distinction between different classes is non-trivial, as for IMDB, or in a multi-class problem as ENZYMES. In general, a 1\% augmentation of the initial dataset is enough to obtain a significant improvement in terms of performance. It is important to remark that, in most cases, graphons estimated using the Gromow-Wasserstein Barycenters lead to better performances.

\section{Conclusion}\label{sec13}
Data from many systems across various areas can be explicitly denoted
as graphs such as social networks, transportation networks, protein–protein
interaction networks. Deep learning approaches have a substantial role in analyzing networks; while many data augmentation techniques exist for numerical, image or text data, this field is still under-explored on graphs. Most of the existing methods works by addition and ablation operations on the existing graphs, which can be limiting. A more effective approach consists in learning a generative model of graphs.\\
The goal of this paper is to introduce a new data augmentation framework that rely on graphons estimated using the Gromow-Wasserstein Barycenters. As shown by the previously reported results, such approach can enhance the performance of Machine Learning models. In addition, since graphs are non-Euclidean object, using a non-Euclidean distance (i.e., the Gromow-Wasserstein distance) to estimate the graphons generally brings to better performance. In this context, graphs are seen as discrete probability distributions supported by their nodes equipped with the nodes degree distribution.\\
Recently, other generative methods for graphs have been proposed mainly based on diffusion models \cite{fan2023generative, jo2022score}. These models could fit well in the proposed framework but it is important to remark that the computational complexity of training diffusion models is much higher than estimating graphons.

\backmatter

\bibliographystyle{acm}
\bibliography{sn-bibliography}

\end{document}